%% file: acl2023.tex
\pdfoutput=1
\documentclass[11pt]{article}

\usepackage[]{ACL2023}

\usepackage{times}
\usepackage{latexsym}

\usepackage[T1]{fontenc}
\usepackage[utf8]{inputenc}

\usepackage{microtype}
\usepackage{enumitem}

\usepackage{inconsolata}
\usepackage{xspace}
\usepackage{times}
\usepackage{latexsym}
\usepackage{amsmath}
\usepackage{multicol}
\usepackage{multirow}
\usepackage{graphicx}
\usepackage{longtable}
\usepackage{tabularx}
\usepackage{balance}
\usepackage{booktabs}
\usepackage{tcolorbox}
\usepackage{xcolor}
\usepackage{float}
\usepackage{enumitem}
\usepackage{verbatim}
\usepackage{listings}
\usepackage{array}

\newcommand{\eg}{\textit{e.g.}\xspace}
\newcommand{\ie}{\textit{i.e.}\xspace}

\newcommand{\ourmethod}{EfficientRAG}

\title{EfficientRAG: Efficient Retriever for Multi-Hop Question Answering}

\author{
    Ziyuan Zhuang\textsuperscript{1\footnotemark[1]},
    Zhiyang Zhang\textsuperscript{1\footnotemark[1]},
    Sitao Cheng\textsuperscript{1}, 
    Fangkai Yang\textsuperscript{2\footnotemark[3]}, 
    Jia Liu\textsuperscript{1}, \\
    \textbf{
        Shujian Huang\textsuperscript{1}, 
        Qingwei Lin\textsuperscript{2},
        Saravan Rajmohan\textsuperscript{2},
        Dongmei Zhang\textsuperscript{2},
        Qi Zhang\textsuperscript{2}
    } \\
\textsuperscript{1} State Key Laboratory for Novel Software Technology, Nanjing University 
\textsuperscript{2} Microsoft \\
ziyuan.zhuang@smail.nju.edu.cn
}

\begin{document}
\maketitle

\renewcommand{\thefootnote}{\fnsymbol{footnote}}
\footnotetext[1]{Equal Contribution}
\footnotetext[2]{Work is done during an internship at Microsoft}
\footnotetext[3]{Corresponding author}
% \footnotetext[3]{Corresponding author}
\renewcommand{\thefootnote}{\arabic{footnote}}

\begin{abstract}
Retrieval-augmented generation (RAG) methods encounter difficulties when addressing complex questions like multi-hop queries.
While iterative retrieval methods improve performance by gathering additional information, current approaches often rely on multiple calls of large language models (LLMs).
In this paper, we introduce EfficientRAG, an efficient retriever for multi-hop question answering.
EfficientRAG iteratively generates new queries without the need for LLM calls at each iteration and filters out irrelevant information.
Experimental results demonstrate that EfficientRAG surpasses existing RAG methods on three open-domain multi-hop question-answering datasets.
The code is available in \href{https://github.com/NIL-zhuang/EfficientRAG-official}{aka.ms/efficientrag}.
\end{abstract}

\input{section/01introduction}

\input{section/Empirical}

\input{section/03methodology}

\input{section/04Experiments}

\input{section/05Analysis}

\input{section/06Conclusion}

\newpage
\section*{Limitations}
The EfficientRAG framework can theoretically adapt to other models, but we opt not to implement a larger LLM as the final QnA reasoner due to time and resource limits.
We analyze our method mainly on open-domain datasets, as it is hard to identify multi-hop question-answering datasets in in-domain settings.

\section*{Ethics Statement}
The authors declare no competing interests. The datasets used in the training and evaluation come from publicly available sources and do not contain sensitive content such as personal information.

\bibliography{custom}
\bibliographystyle{acl_natbib}

\appendix

\input{section/Appendix}

\end{document}

%% file: section/01introduction.tex
\section{Introduction}

Large-language models (LLMs) have shown remarkable performance in numerous applications and tasks~\cite{DBLP:journals/corr/abs-2303-08774, DBLP:journals/corr/abs-2310-06825,DBLP:journals/corr/abs-2307-09288}. However, LLMs lack knowledge underrepresented in their training data, especially in domain-specific settings, and still face the issues of hallucinations~\cite{DBLP:journals/corr/abs-2309-01219, DBLP:journals/corr/abs-2311-05232,DBLP:conf/emnlp/Yang00W0ZGLRZ23}. Retrieval-augmented generation (RAG) techniques~\cite{DBLP:conf/nips/LewisPPPKGKLYR020, DBLP:journals/corr/abs-2312-10997} have been widely adapted to retrieve knowledge from external resources to ground the generated responses. Previous RAG methods often adapt one-round retrieval, \eg, only use the user query or question as the input to retrieve knowledge~\cite{DBLP:conf/icml/GuuLTPC20, DBLP:conf/icml/BorgeaudMHCRM0L22, DBLP:journals/jmlr/IzacardLLHPSDJRG23, DBLP:journals/corr/abs-2301-12652}. Such one-round RAG is capable of answering questions which clearly state all the needed information in the input query~\cite{DBLP:conf/naacl/ThorneVCM18,DBLP:conf/rep4nlp/TrischlerWYHSBS17,DBLP:conf/emnlp/RajpurkarZLL16}, such as one-hop question, \eg, \textit{``what is Newton's third law of motion?''}. However, one-round RAG methods could fail in complex questions where more information is required beyond the first-round retrieved information, \eg, multi-hop questions~\cite{DBLP:conf/emnlp/Yang0ZBCSM18,DBLP:journals/tacl/TrivediBKS22,DBLP:conf/coling/HoNSA20}. 
In order to deal with complex multi-hop questions, recent works propose to obtain required information through multi-round retrievals or reasonings, such as rewriting or generating queries for the following multi-round retrievals~\cite{DBLP:journals/corr/abs-2212-14024,DBLP:journals/corr/abs-2305-14283,DBLP:conf/emnlp/ShaoGSHDC23, DBLP:conf/emnlp/JiangXGSLDYCN23}, interleaving multiple retrieval and reasoning steps~\cite{DBLP:conf/acl/TrivediBKS23}, multi-rounds of self-asking~\cite{DBLP:conf/emnlp/PressZMSSL23}. However, such iterative retrieval approaches have the following limitations: (1) they require multiple LLM calls concerning rewriting or generating new queries for the next round of retrieval, thus increasing the latency and cost. (2) they require dedicated prompting and few-shot examples that might need updating across different scenarios. 

In this paper, we are inspired by the intuition that the types of relations in multi-hop questions are limited, or significantly fewer compared to the number of entities. As proved in \citet{zhu_mirror_2023} that small models have a certain ability of reasoning, we propose that identifying relations and their associated entities from retrieved information can be effectively managed by small models instead of LLMs. Thus, we propose EfficientRAG consists of a Labeler and a Filter to iteratively generate new queries for retrieval and in the meanwhile keep the most relevant retrieved information, enhancing efficiency compared to other RAG methods.

%% file: section/Empirical.tex
\section{Empirical Study}

\subsection{Capability of LLM generator}
\label{sec:capability}
\begin{figure}[htbp]
    \centering
    \begin{minipage}{0.4\textwidth}
        \includegraphics[width=\linewidth]{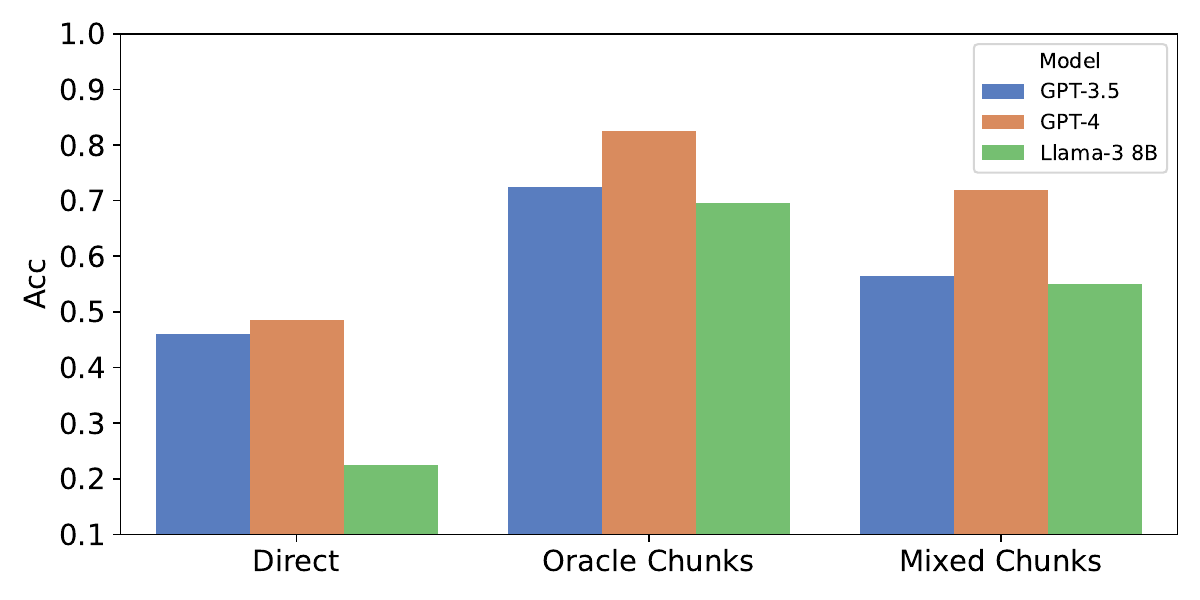}
        \caption{The performance with varying chunks settings over 2WikiMQA dataset with GPT-3.5/GPT-4/Llama3-8B as the generator.}
        \label{fig:llm_generator}
    \end{minipage}
    \vspace{-4mm}
\end{figure}

In this section, we conducted an empirical study to assess how well an LLM-based generator performs with different levels of retrieved information.
We test on three settings: direct prompt (no retrieved chunks), oracle chunks (oracle chunks as the context), and mixed chunks (both oracle and irrelevant chunks as the context) on three datasets, \ie, HotpotQA~\cite{yang2018hotpotqa}, 2Wiki-multihop (2WikiMQA)~\cite{xanh2020_2wikimultihop} and MuSiQue~\cite{trivedi2021musique}.
The generator model includes GPT-3.5~\cite{gpt35}, GPT-4~\cite{DBLP:journals/corr/abs-2303-08774} with 1106-preview version, and Llama-3-8B\footnote{https://llama.meta.com/llama3/}~\cite{DBLP:journals/corr/abs-2302-13971}.
We evaluate the model answer with accuracy metric by GPT-3.5, the prompt can be found in Appendix~\ref{apdx:acc_evaluation}.
As illustrated in Figure ~\ref{fig:llm_generator}, retrieval proves beneficial, with both oracle and mixture settings outperforming the direct answering approach. Nonetheless, the presence of irrelevant chunks continues to challenge the LLM generator, underscoring the need for more precise information retrieval.

\subsection{Retrieve with Query Decomposition}
\begin{figure}[htbp]
    \centering
    \begin{minipage}{0.45\textwidth}
        \includegraphics[width=\linewidth]{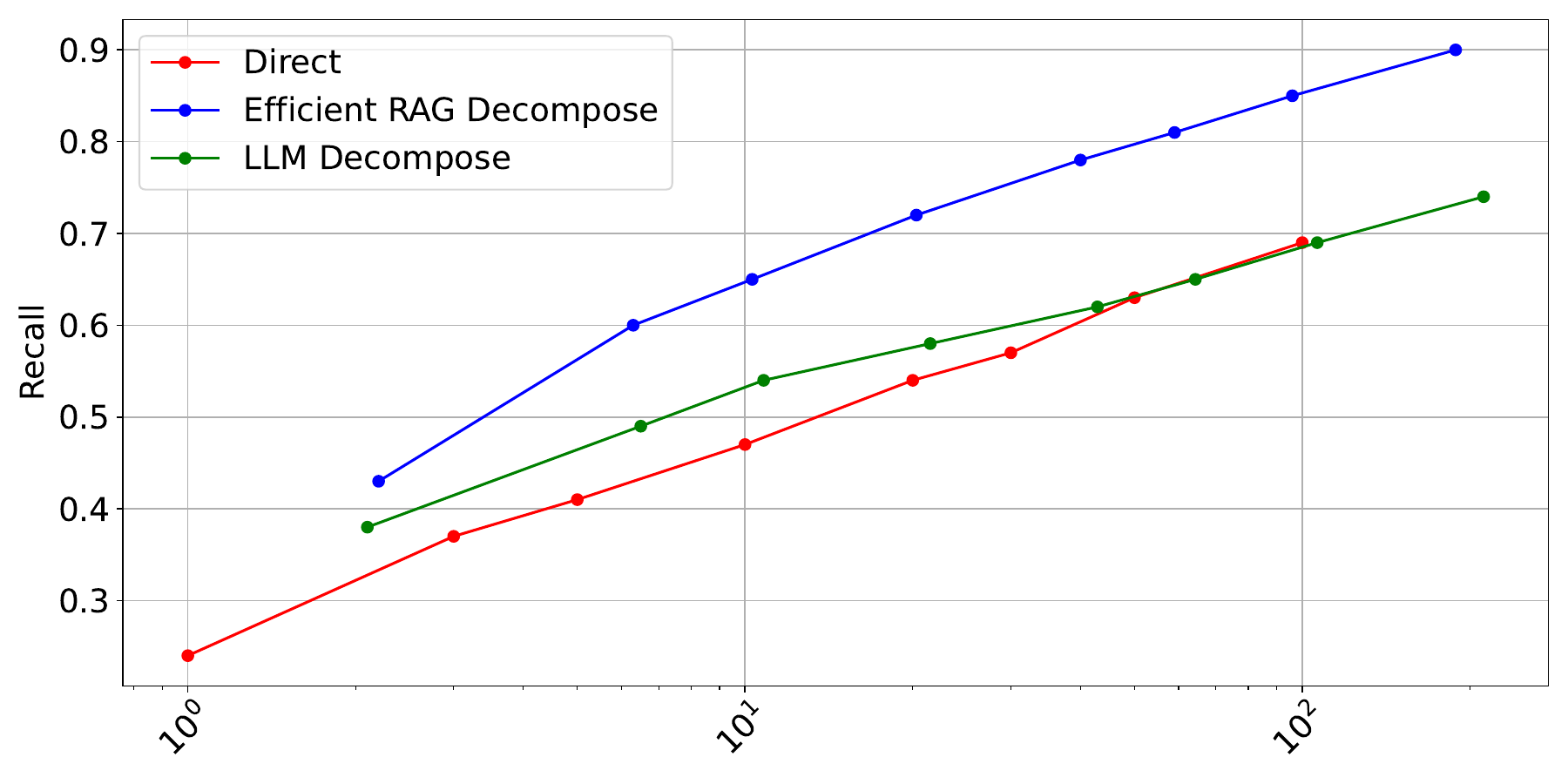}
        \caption{Recall of retrieve efficiency over three retrieval strategies on MuSiQue dataset. The x-axis is log-scaled. Each point on different lines represents the same number of retrieved chunks.}
        \label{fig:retrieve_efficiency}
    \end{minipage}
    \vspace{-4mm}
\end{figure}

It is a common practice to use LLMs for query decomposition when facing complex multi-hop questions~\cite{DBLP:journals/corr/abs-2312-10997}. We conduct another empirical study to check how query decomposition approaches impact the retrieval stage. As shown in Figure~\ref{fig:retrieve_efficiency}, the number of oracle chunks retrieved by one-time decomposition (LLM Decompose, detailed in Table \ref{tab:decompose-prompt}) outperforms the Direct retrieval for the original query. 
At a similar number of chunks, iterative decomposition (EfficientRAG Decompose) achieves higher recall. 
When retrieving approximately 20 chunks, the Recall achieved by EfficientRAG Decompose has comparable performance with the LLM Decompose when retrieving around 200 chunks, thus demonstrating the efficiency of EfficientRAG Decompose. All retrievers used the contriever-msmarco~\cite{DBLP:journals/tmlr/IzacardCHRBJG22} setup, with chunk retrievals configured as 1/3/5/10/20/30/50/100, and the LLM endpoint is gpt35-turbo-1106.

%% file: section/03methodology.tex
\section{Methodology}

\begin{figure*}[tb]
\centering
     \includegraphics[width=0.9\textwidth]{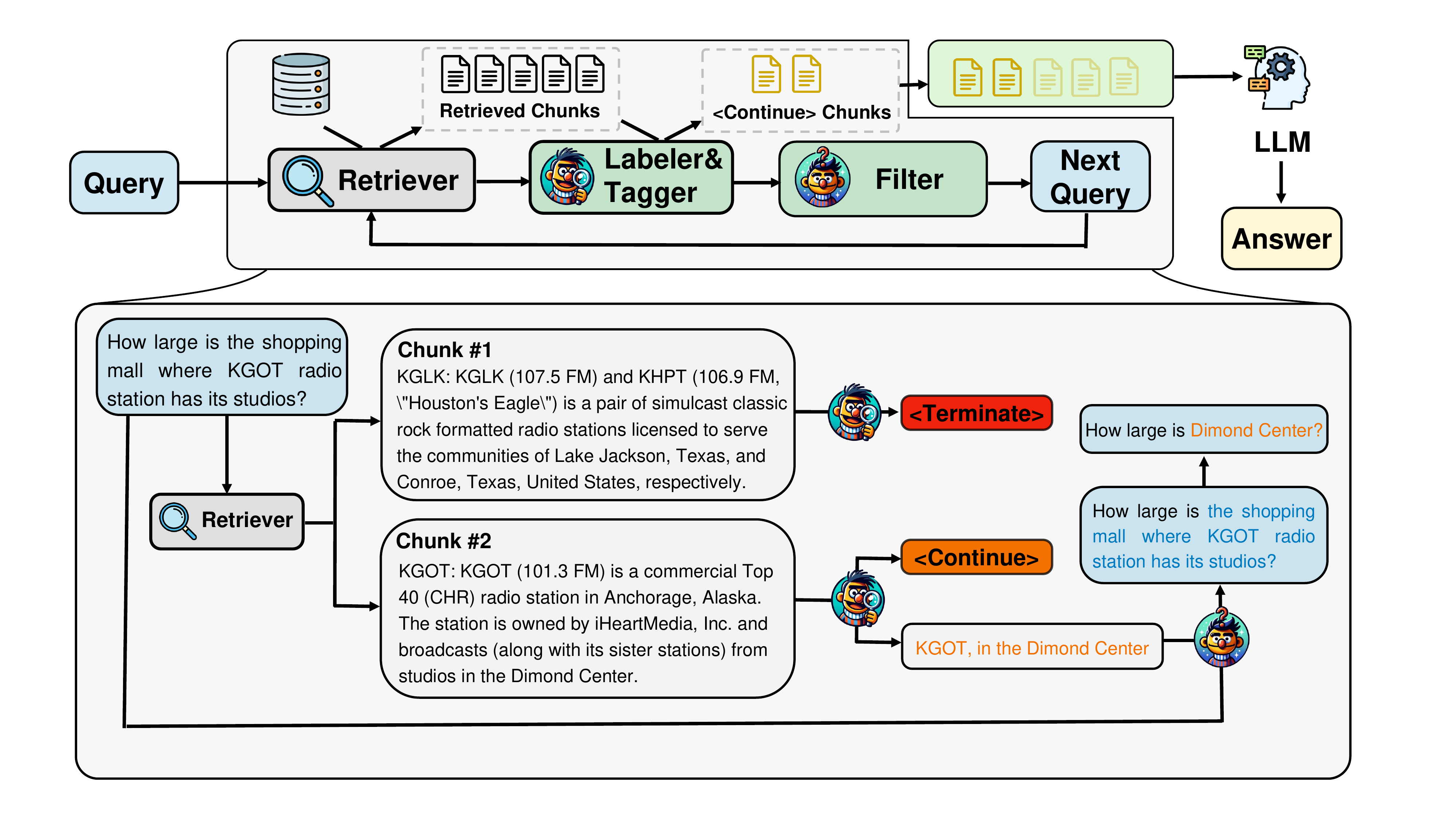}
     \caption{\ourmethod~framework operates within the iterative RAG system. Initially, \ourmethod~retrieves relevant chunks from the knowledge base, tagging each as either <Terminate> or <Continue>, and annotates preserved tokens \textit{"KGOT in the Dimond Center"} from the <Continue> chunks. The Filter then processes the concatenation of the original question and the previously annotated tokens, \textit{"Q: How large is the shopping mall where KGOT radio station has its studios? Info: KGOT, in the Dimond Center"}, and annotates the next-hop query tokens \textit{"How large is Dimond Center?"}. This iterative process continues until all chunks are tagged <Terminate> or the maximum number of iterations is reached.}
     \label{fig:framework}
     \vspace{-4mm}
\end{figure*}

\subsection{\ourmethod~Framework}

In this section, we introduce \ourmethod~, a plug-and-play approach designed to efficiently retrieve relevant information with multiple retrieval rounds to enrich the retrieved information and reduce irrelevant information, then help improve the quality and accuracy of answers. 

\ourmethod~ consists of two lightweight components: the Labeler \& Tagger and the Filter.
These components share the same model structure, with the Labeler \& Tagger\footnote{We use the term Labeler as the representation.} producing outputs from separate heads within the same model and the filter's output comes from another model.
Both the Labeler and the Filter function as token-level classifiers, classifying tokens as either true or false.
Figure~\ref{fig:framework} shows that how \ourmethod~ fits into traditional RAG systems.
Given a query, the retriever obtains relevant chunks from the database.
Then the labeler module annotates a sequence of tokens in this document representing the useful information that could (partially) answer the query.
The tagger module then tags the chunk, indicating whether the retrieved chunk is helpful or irrelevant. If the tag indicates there needs more information to answer the query, \ie, tagged as <Continue>, we will add this chunk to a candidate pool, which will be fed to the LLM-based generator to have the final answer.
Otherwise, if the document is labeled useless or irrelevant, we stop searching for the successor branches from this query.
The filter module takes both the labeled tokens and the current query to construct a new query for the next round of retrieval.
It is done by replacing the unknown part of the query with labeled tokens (useful information). 

Our approach efficiently generates new queries for subsequent retrieval rounds, aiming to retrieve information beyond the scope of the initial query. Once our approach gets enough information to answer the initial question, it stops and passes all this information to the final generator to get the final response. Leveraging our efficient RAG approach eliminates the need for multiple LLM calls for query generation, resulting in improved performance when tackling complex queries.

\subsection{Synthetic Data Construction}
\label{sec:data-construction}
We utilize LLM to synthesize training data for the Labeler and Filter.
The process consists of multi-hop question decomposition, token labeling, next-hop question filtering, and negative sampling.
Synthetic data is detailed in Table \ref{tab:training_data}.

\noindent\textbf{\textit{Multi-hop question decomposition.}} Given a multi-hop question and relevant chunks, we first prompt the LLM to decompose the original question into several single-hop questions. Each single-hop question corresponds to a chunk. Then, we ask the LLM to parse the dependency for the sub-questions.

\noindent\textbf{\textit{Token Labeling.}} For each sub-question and corresponding chunk, we prompt the LLM to label important words in the chunk pertinent to the sub-question answering. We annotate each word in the chunk with a binary label to determine if it is important and should be preserved by EfficientRAG Labeler.
We use the SpaCy toolkit\footnote{https://spacy.io/} following ~\citet{DBLP:journals/corr/abs-2403-12968}.

\noindent\textbf{\textit{Next-hop question filtering.}} Given a single-hop question and the labeled tokens from its dependent questions, we prompt the LLM to generate a next-hop question, which is ideally the next query for retrieval. We extract the next-hop question tokens same as the Token Labeling procedure.

\noindent\textbf{\textit{Negative Sampling.}} With each filtered next-hop question, we retrieve the most similar but not relevant chunk as the hard negative chunk.
These negative chunks will be tagged <Terminate> while other relevant chunks are tagged <Continue>.

\begin{table}[htbp]
\centering
\footnotesize
\begin{tabular}{lcccc}
    \toprule
    Dataset & HotpotQA & MuSiQue & 2WikiMQA \\
    \midrule
    Labeler & 357k & 93k & 70k &  \\
    Filter  &  73k & 25k & 13k &  \\
    \bottomrule
\end{tabular}
\caption{Amount of synthesized training data for different datasets.}
\label{tab:training_data}
\vspace{-4mm}
\end{table}

\begin{table*}[!tb]
\centering
\footnotesize
\begin{tabular}{l|cc|cc|cc}
    \toprule
    \textbf{Method/Dataset} & \multicolumn{2}{c|}{\textbf{HotpotQA}} & \multicolumn{2}{c|}{\textbf{MuSiQue}} & \multicolumn{2}{c}{\textbf{2WikiMQA}} \\ \hline
    \textbf{} & \textbf{Recall@K} & \textbf{K} & \textbf{Recall@K} & \textbf{K} & \textbf{Recall@K} & \textbf{K} \\ \hline
    \textbf{Direct-R@10}            & 70.52 & \underline{10.00} & 50.86 & \underline{10.00} & 61.58 & \underline{10.00} \\
    \textbf{Direct-R@20}            & 74.87 & 20.00 & 57.12 & 20.00 & 65.24 & 20.00 \\
    \textbf{Direct-R@30}            & 77.05 & 30.00 & \underline{60.67} & 30.00 & 66.91 & 30.00 \\
    \textbf{Decompose-R}            & 74.38 & 21.31 & \textbf{67.23} & 19.74 & 71.02 & 21.24 \\
    \textbf{Iter-RetGen iter2} & 81.69 & 14.44 & 57.74 & 14.82 & 73.80 & 14.95 \\
    \textbf{Iter-RetGen iter3} & \textbf{83.05} & 16.42 & 58.26 & 17.19 & 74.29 & 17.32 \\
    \textbf{SelfASK}           & 73.42 & 35.27 & 60.43 & 32.36 & \textbf{88.90} & 33.68 \\
    \textbf{EfficientRAG}           & \underline{81.84} &  \textbf{6.41} & 49.51 & \textbf{6.09} & \underline{84.08} & \textbf{3.69} \\ \bottomrule
\end{tabular}
\caption{Results on retrieval performance. Baselines are implemented from the source code.  Bold and underlined fonts denote the best and second-best results respectively. EfficientRAG demonstrates comparable recall while retrieving the fewest number of chunks.}
\label{tab:retrievalperformance}
\end{table*}

\begin{table*}[ht]
\centering
\footnotesize
\begin{tabular}{l|ccc|ccc|ccc}
    \toprule
    \multicolumn{1}{c|}{\textbf{Method/Dataset}} & \multicolumn{3}{c|}{\textbf{HotpotQA}} & \multicolumn{3}{c|}{\textbf{MuSiQue}} & \multicolumn{3}{c}{\textbf{2WikiMQA}} \\ \hline
    \multicolumn{1}{c|}{\textbf{}} & \multicolumn{1}{c|}{\textbf{EM}} & \multicolumn{1}{c|}{\textbf{F1}} & \textbf{Acc} & 
                                     \multicolumn{1}{c|}{\textbf{EM}} & \multicolumn{1}{c|}{\textbf{F1}} & \textbf{Acc} & 
                                     \multicolumn{1}{c|}{\textbf{EM}} & \multicolumn{1}{c|}{\textbf{F1}} & \textbf{Acc} \\ \hline
                                     
    \textbf{Direct} & \multicolumn{1}{c|}{22.87} & \multicolumn{1}{c|}{26.94} & \multicolumn{1}{c|}{25.79} & \multicolumn{1}{c|}{5.59} & \multicolumn{1}{c|}{8.76} & \multicolumn{1}{c|}{5.51} & \multicolumn{1}{c|}{27.33} & \multicolumn{1}{c|}{31.11} & \multicolumn{1}{c}{28.67} \\
    
    \textbf{CoT} & \multicolumn{1}{c|}{27.99} & \multicolumn{1}{c|}{34.05} & \multicolumn{1}{c|}{30.53} & \multicolumn{1}{c|}{10.16} & \multicolumn{1}{c|}{13.85} & \multicolumn{1}{c|}{9.21} & \multicolumn{1}{c|}{29.25} & \multicolumn{1}{c|}{35.14} & \multicolumn{1}{c}{31.71} \\
    
    \textbf{Direct-R@10} & \multicolumn{1}{c|}{38.24} & \multicolumn{1}{c|}{44.55} & \multicolumn{1}{c|}{44.56} & \multicolumn{1}{c|}{13.39} & \multicolumn{1}{c|}{18.14} & \multicolumn{1}{c|}{17.12} & \multicolumn{1}{c|}{26.18} & \multicolumn{1}{c|}{31.88} & \multicolumn{1}{c}{32.70} \\
    
    \textbf{Decompose-R} & \multicolumn{1}{c|}{36.15} & \multicolumn{1}{c|}{42.68} & \multicolumn{1}{c|}{46.31} & \multicolumn{1}{c|}{12.59} & \multicolumn{1}{c|}{19.25} & \multicolumn{1}{c|}{19.53} & \multicolumn{1}{c|}{25.22} & \multicolumn{1}{c|}{31.44} & \multicolumn{1}{c}{32.53} \\
    
    \textbf{Iter-RetGen iter2} & \multicolumn{1}{c|}{55.45} & \multicolumn{1}{c|}{59.47} & \multicolumn{1}{c|}{\textbf{59.29}} & \multicolumn{1}{c|}{26.95} & \multicolumn{1}{c|}{29.15} & \multicolumn{1}{c|}{\underline{26.28}} & \multicolumn{1}{c|}{43.85} & \multicolumn{1}{c|}{49.96} & \multicolumn{1}{c}{49.22} \\
    
    \textbf{Iter-RetGen iter3} & \multicolumn{1}{c|}{56.76} & \multicolumn{1}{c|}{60.89} & \multicolumn{1}{c|}{57.56} & \multicolumn{1}{c|}{28.20} & \multicolumn{1}{c|}{30.31} & \multicolumn{1}{c|}{25.31} & \multicolumn{1}{c|}{43.07} & \multicolumn{1}{c|}{49.83} & \multicolumn{1}{c}{46.59} \\
    
    \textbf{SelfASK} & \multicolumn{1}{c|}{33.58} & \multicolumn{1}{c|}{39.10} & \multicolumn{1}{c|}{42.36} & \multicolumn{1}{c|}{24.56} & \multicolumn{1}{c|}{29.22} & \multicolumn{1}{c|}{\textbf{26.97}} & \multicolumn{1}{c|}{47.56} & \multicolumn{1}{c|}{54.84} & \multicolumn{1}{c}{\textbf{55.16}} \\
    
    \textbf{EfficientRAG} & \multicolumn{1}{c|}{50.59} & \multicolumn{1}{c|}{57.93} & \multicolumn{1}{c|}{\underline{57.86}} & \multicolumn{1}{c|}{16.44} & \multicolumn{1}{c|}{21.18} & \multicolumn{1}{c|}{20.00} & \multicolumn{1}{c|}{44.18} & \multicolumn{1}{c|}{51.64} & \multicolumn{1}{c}{\underline{53.41}} \\ \bottomrule
\end{tabular}
\caption{Results on end-to-end question answering performance across three datasets. The highest accuracy (Acc) values are highlighted in bold, while the second-highest are underlined. EfficientRAG exhibits promising high accuracy, comparable to that of the LLM-based baselines.}
\label{tab:QAperformance}
\vspace{-4mm}
\end{table*}

\subsection{Training}
We train EfficientRAG Labeler for two tasks, token labeling and chunk filtering, as they both take in the same input.
We use an auto-encoder language model as an encoder to derive embeddings for each token of concatenated sequence $\text{query}, \text{chunk}$.
Subsequently, we use one fully connected layer to project the token embedding into a 2-dimensional space, indicating "useful token" and "useless token".
Another fully connected layer is adapted to project the average pooling of the sequence embedding into a 2-dimensional space, representing the chunk tag <Continue> and <Terminate>.
We train \ourmethod Filter similarly, while its input sequence is the concatenation of query and labeled tokens.
The Filter extracts words and concatenates them to formulate the next-hop query.

%% file: section/04Experiments.tex
\section{Experiments}

\subsection{End2end QA performance}
We conduct evaluations of our EfficientRAG and multiple baselines on three multi-hop question-answering datasets same as \S\ref{sec:capability}.
We select the following models as our baselines.
First is direct answering without retrieval, including LMs with proprietary data.
We include direct prompting and Chain-of-Thought prompting~\cite{DBLP:journals/corr/abs-2302-13971} and question decomposition prompting in this setting.
Secondly, we include baselines with naive RAG with top-10 retrieve chunks as its knowledge.
Third, we include advanced iterative RAG methods like Iter-RetGen~\cite{DBLP:conf/emnlp/ShaoGSHDC23} and SelfAsk~\cite{DBLP:conf/emnlp/PressZMSSL23}.
The implementation prompts are in Appendix ~\ref{apdx:baseline_implementation}.

\noindent \textit{\textbf{Implementation Details.}}
EfficientRAG Labeler and Filter are fine-tuned based on DeBERTa-v3-large~\cite{DBLP:conf/iclr/HeLGC21} with 24 layers and 304M parameters. We adopt Llama-3-8B-Instruct for the question-answering stage and all other baselines. We utilize Contriever-MSMARCO~\cite{DBLP:journals/tmlr/IzacardCHRBJG22} as the retriever for both data synthesis and inference stages.

We constructed the training data following Section \ref{sec:data-construction} with Llama-3-70B-Instruct (Prompts are detailed in Appendix ~\ref{apdx:data_synthesize_prompt}).
We trained our model on $4\times$ Nvidia A100 GPUs for about 10 GPU-hours separately, with AdamW~\cite{DBLP:conf/iclr/LoshchilovH19} optimizer and a learning rate of 5e-6.

%% file: section/05Analysis.tex
\section{Results and Analysis}

\subsection{Retrieval Performance}

The model's performance was assessed using the Recall@K metric across three distinct datasets. As presented in Table ~\ref{tab:retrievalperformance}, EfficientRAG achieves notably high recall scores on HotpotQA and 2WikiMQA datasets, with recall values of 81.84 and 84.08, respectively. These results are impressive considering the minimal number of chunks retrieved 6.41 for HotpotQA and 3.69 for 2WikiMQA. However, the performance of EfficientRAG on the MuSiQue dataset was less satisfactory. This suboptimal result may be attributed to the smaller number of chunks retrieved and the increased complexity of the dataset.

We further evaluate the QnA performance on the three datasets.
As is illustrated in Table ~\ref{tab:QAperformance}, our EfficientRAG framework achieves the second-highest accuracy on both HotpotQA and 2WikiMQA, and it also performs well on MuSiQue even with low recall.

Those LLM-based systems perform unsatisfying since they require LLMs to generate partial answers with noisy knowledge inputs, but they always fail in the intermediate steps.
We posit that more helpful knowledge and fewer irrelevant chunks are the key points to the RAG system, even a simple model can beat LLMs with the correct RAG paradigm.

\subsection{Inference Efficiency}
We randomly selected 200 samples from the MusiQue dataset for empirical research and calculated four indicators: LLM calls, iterations, latency, and GPU utilization. As shown in table \ref{tab:efficiency} our method requires fewer iterations and achieves a 60\%-80\% improvement in time efficiency compared to other iterative methods while maintaining similar GPU utilization.

\begin{table}[ht]
    \footnotesize
    \centering
    \resizebox{0.49\textwidth}{!}{
    \begin{tabular}{l|c|c|c|c}
        \toprule
        \multicolumn{1}{c|}{\textbf{Method}} & \textbf{\# LLM calls} & \textbf{\# Iteration} & \textbf{Latency (s)} & \textbf{GPU utils (\%)}\\ \hline
        \textbf{Direct}                      & 1.00 & -     & 2.16 & 15.55 \\
        \textbf{Direct-R}                    & 1.00 & -     & 2.47 & 35.05 \\
        \textbf{Iter-RetGen iter3}           & 3.00 & 3.00  & 9.68 & 66.37 \\
        \textbf{SelfASK}                     & 7.18 & 3.59 & 27.47 & 65.02 \\
        \textbf{EfficientRAG}                & 1.00 & 2.73 & 3.62 & 65.55 \\ \bottomrule
        \end{tabular}
    }
    \caption{Efficiency evaluation on different RAG paradigms. EfficientRAG exhibits a speed equivalent to direct retrieval methods and is three times faster than LLM-based baselines while maintaining a similar number of iterations.}
    \label{tab:efficiency}
\end{table}

\subsection{Performance with Various Generators}

EfficientRAG can benefit from more powerful generators.
As is shown in Table \ref{tab:gpt35-generator}, the use of GPT-3.5 as a generator enhances the end-to-end performance of both the baselines and our method.
Notably, EfficientRAG continues to deliver exceptional results.

\begin{table}[h!]
    \footnotesize
    \centering
    \begin{tabular}{c|c|c|c}
        \toprule
        \multicolumn{1}{c}{\textbf{Method}} & \textbf{EM} & \textbf{F1} & \textbf{Acc} \\
        \hline
        \textbf{Direct}            & 27.85       & 33.22       & 33.79        \\
        \textbf{CoT}               & 38.56       & 47.28       & 46.62        \\
        \textbf{Direct-R}          & 34.09       & 39.46       & 41.07        \\
        \textbf{iter-RetGen iter2} & 47.51       & 58.56       & 58.34        \\
        \textbf{Iter-RetGen iter3} & 49.41       & 60.60       & 60.60        \\
        \textbf{EfficientRAG}      & 49.00       & 56.93       & \textbf{61.88}     \\  
        \bottomrule
    \end{tabular}
    \caption{End-to-end QA performance on the 2WikiMQA dataset using GPT-3.5-turbo-1106 generator. EfficientRAG achieves state-of-the-art accuracy.}
    \label{tab:gpt35-generator}
\end{table}

\subsection{Out-Of-Domain Adaptation}

EfficientRAG has the potential to adapt to different scenarios without further downstream training.
We propose an out-of-domain experiment across HotpotQA and 2WikiMQA datasets, where we train the model on one dataset and test it on the other.
Table \ref{tab:ood} shows that our model adapts well to different datasets, and even surpasses the model trained on the original data in some cases.
It shows that EfficientRAG does not rely on domain-specific knowledge, exhibiting a certain degree of transferability.

\begin{table}[h!]
\footnotesize
\centering
\begin{tabular}{ccccc}
    \toprule
    \textbf{Test Set} & \textbf{Training Set} & \textbf{EM} & \textbf{F1} & \textbf{Acc} \\
    \hline
    HotpotQA & HotpotQA     & 50.59 & 57.93 & 57.86 \\
    HotpotQA & 2WikiMQA     & 43.38 & 49.70 & 53.38 \\
    2WikiMQA & 2WikiMQA     & 44.18 & 51.64 & 53.41 \\
    2WikiMQA & HotpotQA     & 44.54 & 51.98 & 56.59 \\
    \bottomrule
\end{tabular}
\caption{Out-of-domain experiments on 2WikiMQA and HotpotQA dataset. EfficientRAG demonstrates remarkable transferability across diverse datasets.}
\label{tab:ood}
\end{table}

%% file: section/06Conclusion.tex
\section{Conclusion}

In this study, we introduce the EfficientRAG retriever, a novel approach for multi-hop question retrieval that iteratively generates new queries while circumventing the need for large language models.
Evaluations across three benchmark datasets demonstrate that EfficientRAG not only achieves high recall with a minimal number of retrieved chunks but also delivers promising outcomes in subsequent question-answering tasks. 
These findings indicate that EfficientRAG outperforms traditional retrieval-augmented generation methods, particularly in the context of complex, multi-hop question-answering scenarios.

%% file: section/Appendix.tex
\newpage

\section{Capability of LLM generator}

\begin{figure}[htbp]
    \centering
    \begin{minipage}{0.45\textwidth}
        \includegraphics[width=\linewidth]{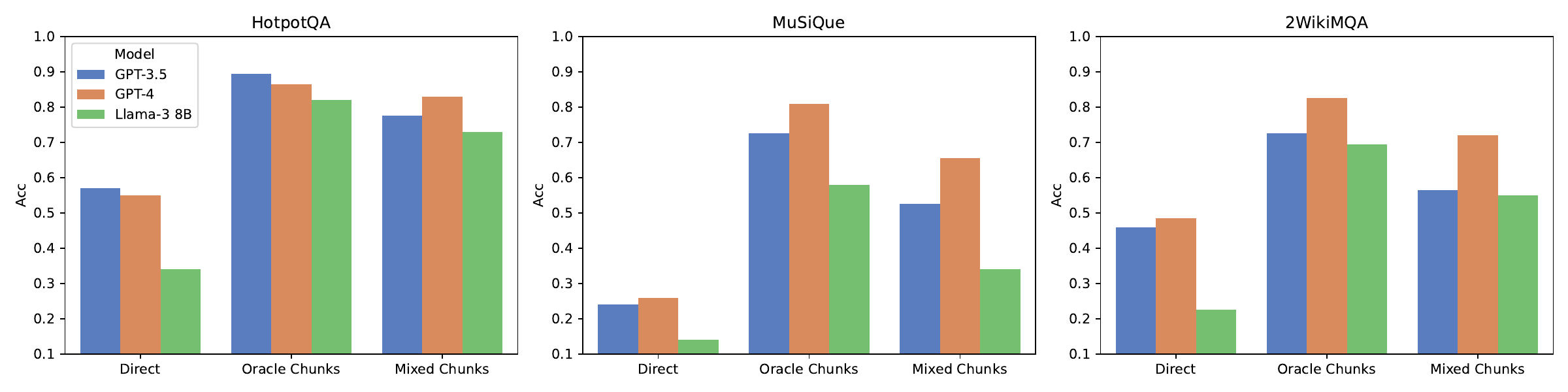}
        \caption{The performance with varying chunks settings over HotpotQA, 2Wiki-Multihop and MuSiQue dataset with GPT-3.5/GPT-4/Llama-3 8B as the generator.}
        \label{fig:llm_generator_full}
    \end{minipage}
\end{figure}

\section{Prompt List}

\subsection{Accuracy Evaluation Prompt}
\label{apdx:acc_evaluation}

\begin{table}[!thb]
    \footnotesize
    \begin{tabular}{p{7 cm}}
    \toprule
    You are an experienced linguist who is responsible for evaluating the correctness of the generated responses.
    You are provided with question, the generated responses and the corresponding ground truth answer.
    Your task is to compare the generated responses with the ground truth responses and evaluate the correctness of the generated responses.
    Response in JSON format with key "response" and value "yes" or "no". \\
    Question: \{question\} \\
    Prediction: \{prediction\} \\
    Ground-truth Answer: \{answer\} \\
    Your response: \\
    \bottomrule
    \end{tabular}
    \caption{Detailed prompts for GPT-3.5 evaluation.}
    \label{tab:acc_prompt}
\end{table}

\onecolumn
\subsection{Data Synthesize Prompt}
\label{apdx:data_synthesize_prompt}

\begin{table*}[htbp]
    \setlength{\abovecaptionskip}{0.1cm}
    \setlength{\belowcaptionskip}{-0.5cm}
    \centering
    \footnotesize
    \begin{tabular}{p{15cm}}
    \toprule
    
    \textbf{Question Decomposition Prompt} \\
    You are assigned a multi-hop question decomposition task. \\
    Your mission is to decompose the original multi-hop question into a list of single-hop sub\_questions, based on supporting documents for each sub\_question, and such that you can answer each sub\_question independently from each document. Each document infers a sub\_question id which starts with `\#`. The evidence in the document indicates the relation of two entities, in the form of `entity1 - relation - entity2`. \\
    The JSON output must contain the following keys: \\
    - "question": a string, the original multi-hop question. \\
    - "decomposed\_questions": a dict of sub\_questions and answers. The key should be the sub\_question number(string format), and each value should be a dict containing: \\
        - "sub\_question": a string, the decomposed single-hop sub\_question. It MUST NOT contain information more than the original question and its dependencies. NEVER introduce information from documents. \\
        - "answer": a string, the answer of the sub\_question. \\
        - "dependency": a list of sub\_question number(string format). If the sub\_question relies on the answer of other sub\_questions, you should list the sub\_question number here. Leave it empty for now because the questions now are all comparison type. \\
        - "document": a string, the document id that supports the sub\_question. \\
    Notice that you don't need to come out the compare question, just the sub\_questions and answers. \\
    \midrule
    \textbf{Token Labeling Prompt} \\
    You have been assigned an information extraction task. \\
    Your mission is to extract the words from a given paragraph so that others can answer a question using only your extracted words. \\
    Your extracted words should cover information from both the question and the answer, including entities (e.g. people, location, film) and core relations. \\
    Your response should be in JSON format and include the following key: \\
    - "extracted\_words": a string composed of a list of words extracted from the paragraph, separated by a space. 
    
    Please adhere to the following guidelines: \\
    - Do not reorder, change, miss, or add words. Keep it the same as the original paragraph. \\
    - Identify and extract ONLY the words explicitly mentioned in either the question or its answer, and strongly related to the question or its answer. \\
    - NEVER label any words that do not contribute meaningful information to the question or answer. \\
    - Only extract words that occurred in the paragraph. \\
    \midrule
    \textbf{Query Filtering Prompt} \\
    You are assigned a multi-hop question refactoring task. \\
    Given a complex question along with a set of related known information, you are required to refactor the question by applying the principle of retraining difference and removing redundancies. Specifically, you should eliminate the content that is duplicated between the question and the known information, leaving only the parts of the question that have not been answered, and the new knowledge points in the known information. The ultimate goal is to reorganize these retrained parts to form a new question. \\
    You can only generate the question by picking words from the question and known information. You should first pick up words from the question, and then from each known info, and concatenate them finally. You are not allowed to add, change, or reorder words. The given known information starts with the word "Info: ". \\
    You response should be in JSON format and include the following key: \\
    - "filtered\_query": a string representing the concatenation of the words from both the question and newly added information, separated by a space. \\
    
    Please adhere to the following guidelines: \\
    - Do not reorder, change, or add words. Keep it the same as the original question. \\
    - Identify and remove ONLY the words that are already known, keep the unknown information from both the question and information. \\
    \bottomrule
    \end{tabular}
    \caption{Detailed prompts for training data construction with Llama-3 70B}
    \label{tab:data_construction}
\end{table*}

\subsection{Prompts List}

All prompts can be found in this section, and are given in the order of Direct, CoT, Decompose, Direct-R, Iter-RetGen, and Self-ask, as shown in Tables \ref{tab:direct_prompt} to \ref{tab:self-ask_pormpt_2WikiMQA}.
\label{apdx:baseline_implementation}

\begin{table*}[htbp]
    \setlength{\abovecaptionskip}{0.1cm}
    \setlength{\belowcaptionskip}{-0.5cm}
    \centering
    \footnotesize
    \begin{tabular}{p{15cm}}
    \toprule
        \textbf{Direct Prompting for HotpotQA} \\
        As an assistant, your task is to answer the question directly after <Question>. Your answer should be after <Answer> in JSON format with key "answer" and its value should be string. \\

        There are some examples for you to refer to: \\
        <Question>: What is the name of this American musician, singer, actor, comedian, and songwriter, who worked with Modern Records and born in December 5, 1932? \\
        <Answer>: \\
        \texttt{\`}\texttt{\`}\texttt{\`} json \\
        \{\{"answer": "Little Richard"\}\} \\
        \texttt{\`}\texttt{\`}\texttt{\`} \\
        
        <Question>: Between Chinua Achebe and Rachel Carson, who had more diverse jobs? \\
        <Answer>: \\
        \texttt{\`}\texttt{\`}\texttt{\`} json \\
        \{\{"answer": "Chinua Achebe"\}\} \\
        \texttt{\`}\texttt{\`}\texttt{\`} \\
        
        <Question>: Remember Me Ballin' is a CD single by Indo G that features an American rapper born in what year? \\
        <Answer>: \\
        \texttt{\`}\texttt{\`}\texttt{\`} json \\
        \{\{"answer": "1979"\}\} \\
        \texttt{\`}\texttt{\`}\texttt{\`} \\
        
        Now your Question is \\
        <Question>: \{question\} \\
        <Answer>: \\
        \midrule
        \textbf{Direct Prompting for MuSiQue} \\
        As an assistant, your task is to answer the question directly after <Question>.
        Your answer should be after <Answer> in JSON format with key "answer" and its value should be string.
        
        There are some examples for you to refer to: \\
        <Question>: In which year did the publisher of In Cold Blood form? \\
        <Answer>: \\
        \texttt{\`}\texttt{\`}\texttt{\`}json\\
        \{\{"answer": "2001"\}\} \\
        \texttt{\`}\texttt{\`}\texttt{\`} \\
        
        <Question>: Who was in charge of the city where The Killing of a Sacred Deer was filmed? \\
        <Answer>: \\
        \texttt{\`}\texttt{\`}\texttt{\`} json \\
        \{\{"answer": "John Cranley"\}\} \\
        \texttt{\`}\texttt{\`}\texttt{\`}
        
        <Question>: Where on the Avalon Peninsula is the city that Signal Hill overlooks? \\
        <Answer>: \\
        \texttt{\`}\texttt{\`}\texttt{\`}json\\
        \{\{"answer": "eastern tip"\}\} \\
        \texttt{\`}\texttt{\`}\texttt{\`}
        
        Now your Question is \\
        <Question>: \{question\} \\
        <Answer>:  \\
        \midrule
        \textbf{Direct Prompting for 2Wiki-MultihopQA} \\
        As an assistant, your task is to answer the question directly after <Question>. Your answer should be after <Answer> in JSON format with key "answer" and its value should be string.\\
        
        There are some examples for you to refer to:\\
        <Question>: Which film came out first, Blind Shaft or The Mask Of Fu Manchu?\\
        <Answer>:\\
        \texttt{\`}\texttt{\`}\texttt{\`}json\\
        \{\{"answer": "The Mask Of Fu Manchu"\}\}\\
        \texttt{\`}\texttt{\`}\texttt{\`} \\
        
        <Question>: When did John V, Prince Of Anhalt-Zerbst's father die?\\
        <Answer>:\\
        \texttt{\`}\texttt{\`}\texttt{\`}json\\
        \{\{"answer": "12 June 1516"\}\}\\
        \texttt{\`}\texttt{\`}\texttt{\`}\\
        
        <Question>: Which film has the director who was born later, El Extrano Viaje or Love In Pawn?\\
        <Answer>:\\
        \texttt{\`}\texttt{\`}\texttt{\`}json\\
        \{\{"answer": "El Extrano Viaje"\}\}\\
        \texttt{\`}\texttt{\`}\texttt{\`}\\
        
        Now your Question is\\
        <Question>: \{question\}\\
        <Answer>:\\
    \bottomrule
    \end{tabular}
    \caption{Detailed prompts for Direct Question Answering with Llama-3 8B}
    \label{tab:direct_prompt}
\end{table*}

\begin{table*}[htbp]
    \setlength{\abovecaptionskip}{0.1cm}
    \setlength{\belowcaptionskip}{-0.5cm}
    \centering
    \footnotesize
    \begin{tabular}{p{15cm}}
    \toprule
        \textbf{CoT Prompting for HotpotQA} \\
        As an assistant, your task is to answer the question after <Question>. You should first think step by step about the question and give your thought and then answer the <Question>. Your answer should be after <Answer> in JSON format with key "thought" and "answer" and their values should be string. \\

        There are some examples for you to refer to: \\
        <Question>: What is the name of this American musician, singer, actor, comedian, and songwriter, who worked with Modern Records and born in December 5, 1932? \\
        <Answer>: \\
        \texttt{\`}\texttt{\`}\texttt{\`} json \\
        \{\{"thought":"Modern Record is a big R\&B label with artists including Etta James, Joe Houston, Little Richard, Ike, Tina Turner and John Lee Hooker in the 1950s and 1960s. Little Richard is an American musician, signer actor and songwriter, born in December 5 1932. So the answer is Little Richard.","answer": "Little Richard"\}\} \\
        \texttt{\`}\texttt{\`}\texttt{\`} \\
        
        <Question>: Between Chinua Achebe and Rachel Carson, who had more diverse jobs? \\
        <Answer>: \\
        \texttt{\`}\texttt{\`}\texttt{\`} json \\
        \{\{"thought":"Chinua Achebe was a Nigerian novelist, poet, professor, and critic. Rachel Carson was an American marine biologist, author, and conservationist. Chinua Achebe has 4 jobs while Rachel Carson has 3 jobs. So the answer is Chinua Achebe.","answer": "Chinua Achebe"\}\} \\
        \texttt{\`}\texttt{\`}\texttt{\`} \\
        
        <Question>: Remember Me Ballin' is a CD single by Indo G that features an American rapper born in what year? \\
        <Answer>: \\
        \texttt{\`}\texttt{\`}\texttt{\`} json \\
        \{\{"thought":"Remember Me Ballin' is the CD singer by Indo G that features Gangsta Boo, who is named Lola Mitchell, an American rapper born in 1979. So the answer is 1979.","answer": "1979"\}\} \\
        \texttt{\`}\texttt{\`}\texttt{\`} \\
        
        Now your Question is \\
        <Question>: \{question\} \\
        <Answer>: \\
    \bottomrule
    \end{tabular}
    \caption{Detailed prompts for Chain-of-Thought Question Answering with Llama-3 8B On hotpotQA}
    \label{tab:cot_prompt_hotpotQA}
\end{table*}
    
\begin{table*}[htbp]
\setlength{\abovecaptionskip}{0.1cm}
    \setlength{\belowcaptionskip}{-0.5cm}
    \centering
    \footnotesize
    \begin{tabular}{p{15cm}}
    \toprule
    \textbf{Question Decomposition Prompt}
    You are assigned a multi-hop question decomposition task. \\
    You should decompose the given multi-hop question into multiple single-hop questions, and such that you can answer each single-hop question independently. \\
    Your response must be wrapped with \texttt{\`}\texttt{\`}\texttt{\`}json and \texttt{\`}\texttt{\`}\texttt{\`}. \\
    You should answer in JSON format, your answer must contain the following keys: \\
        - "decomposed\_questions": a list of strings, each string is a single-hop question. \\
    
    Here are some examples for your reference: \\
    \#\# Examples \\
    <Multi-hop question>: Which film came out first, The Love Route or Engal Aasan? \\
    Your response: \\
    \texttt{\`}\texttt{\`}\texttt{\`}json \\
    \{\{  "decomposed\_questions": [  "When does the film The Love Route come out?",  "When does the film Engal Aasan come out?"  ]  \}\} \\
    \texttt{\`}\texttt{\`}\texttt{\`} \\
    
    <Multi-hop question>: Where did the spouse of Moderen's composer die? \\
    Your response: \\
    \texttt{\`}\texttt{\`}\texttt{\`}json \\
    \{\{ "decomposed\_questions": [ "Who is Modern's composer?", "Who is the spouse of Carl Nielsen?",  "In what place did Anne Marie Carl-Nielsen die?"  ]  \}\} \\
    \texttt{\`}\texttt{\`}\texttt{\`} \\
    <Multi-hop question>: Where was the director of film The Fascist born? \\
    Your response: \\
    \texttt{\`}\texttt{\`}\texttt{\`}json \\
    \{\{ "decomposed\_questions": [  "Who is the director of film The Fascist?",  "Where was Luciano Salce born?"  ]  \}\} \\
    \texttt{\`}\texttt{\`}\texttt{\`} \\
     \\
    \#\# Now it's your turn: \\
    <Multi-hop question>: \{question\} \\
    Your response: \\
    \bottomrule
    \end{tabular}
    \caption{Detailed prompts for multi-hop question decomposition, applicable to all datasets.}
    \label{tab:decompose-prompt}
\end{table*}

\begin{table*}[htbp]
    \setlength{\abovecaptionskip}{0.1cm}
    \setlength{\belowcaptionskip}{-0.5cm}
    \centering
    \footnotesize
    \begin{tabular}{p{15cm}}
    \toprule       
        \textbf{CoT Prompting for MuSiQue} \\
        As an assistant, your task is to answer the question after <Question>. You should first think step by step about the question and give your thought and then answer the <Question>. Your answer should be after <Answer> in JSON format with key "thought" and "answer" and their values should be string. \\

        There are some examples for you to refer to: \\
        <Question>: In which year did the publisher of In Cold Blood form? \\
        <Answer>: \\
        \texttt{\`}\texttt{\`}\texttt{\`} json \\
        \{\{"thought": "The publisher of In Cold Blood is Random house, which was formed in 2001. So the answer is 2001.", "answer": "2001"\}\} \\
        \texttt{\`}\texttt{\`}\texttt{\`} \\
        
        <Question>: Who was in charge of the city where The Killing of a Sacred Deer was filmed? \\
        <Answer>: \\
        \texttt{\`}\texttt{\`}\texttt{\`} json \\
        \{\{"thought": "The killing of a Scared Deer was filmed in Cincinnati, Ohio, where John Cranley is the mayor. So the answer is John Cranley.", "answer": "John Cranley"\}\} \\
        \texttt{\`}\texttt{\`}\texttt{\`} \\
        
        <Question>: Where on the Avalon Peninsula is the city that Signal Hill overlooks? \\
        <Answer>: \\
        \texttt{\`}\texttt{\`}\texttt{\`} json \\
        \{\{"thought": "Signal Hill overlooks the city St. John's, which is located on the eastern tip of the Avalon Peninsula. So the answer is eastern tip.", "answer": "eastern tip"\}\} \\
        \texttt{\`}\texttt{\`}\texttt{\`} \\
        
        Now your Question is \\
        <Question>: \{question\} \\
        <Answer>: \\
    \bottomrule
    \end{tabular}
    \caption{Detailed prompts for Chain-of-Thought Question Answering with Llama-3 8B on MuSiQue}
    \label{tab:cot_prompt_musique}
\end{table*}

\begin{table*}[htbp]
    \setlength{\abovecaptionskip}{0.1cm}
    \setlength{\belowcaptionskip}{-0.5cm}
    \centering
    \footnotesize
    \begin{tabular}{p{15cm}}
    \toprule
        \textbf{CoT Prompting for 2Wiki-MultihopQA} \\
        As an assistant, your task is to answer the question after <Question>. You should first think step by step about the question and give your thought and then answer the <Question>. Your answer should be after <Answer> in JSON format with key "thought" and "answer" and their values should be string. \\

        There are some examples for you to refer to: \\
        <Question>: Which film came out first, Blind Shaft or The Mask Of Fu Manchu? \\
        <Answer>: \\
        \texttt{\`}\texttt{\`}\texttt{\`} json \\
        \{\{"thought": "Blind Shaft is a 2003 Chinese film, and The Mask Of Fu Manchu is a 1932 American pre-Code adventure film. The Mask Of Fu Manchu came out first. So the answer is The Mask Of Fu Manchu.", "answer": "The Mask Of Fu Manchu"\}\} \\
        \texttt{\`}\texttt{\`}\texttt{\`} \\
        
        <Question>: When did John V, Prince Of Anhalt-Zerbst's father die? \\
        <Answer>: \\
        \texttt{\`}\texttt{\`}\texttt{\`} json \\
        \{\{"thought": "The father of John V, Prince Of Anhalt-Zerbst is Ernest I, Prince of Anhalt-Dessau. He died on 12 June 1516. So the answer is 12 June 1516.", "answer": "12 June 1516"\}\} \\
        \texttt{\`}\texttt{\`}\texttt{\`} \\
        
        <Question>: Which film has the director who was born later, El Extrano Viaje or Love In Pawn? \\
        <Answer>: \\
        \texttt{\`}\texttt{\`}\texttt{\`} json \\
        \{\{"thought": "The director of El Extrano Viaje is Fernando Fernan Gomez, he was born on 29 August 1921. The director of Love In Pawn is Charles Saunders, he was born on 8 April 1904. Fernando Fernan Gomez was born later, so film El Extrano Viaje has the director who was born later. So the answer is El Extrano Viaje.", "answer": "El Extrano Viaje"\}\} \\
        \texttt{\`}\texttt{\`}\texttt{\`} \\
        
        Now your Question is \\
        <Question>: \{question\} \\
        <Answer>: \\
    \bottomrule
    \end{tabular}
    \caption{Detailed prompts for Chain-of-Thought Question Answering with Llama-3 8B on 2Wiki-MultihopQA}
    \label{tab:cot_prompt_2WikiMQA}
\end{table*}

\begin{table*}[htbp]
    \setlength{\abovecaptionskip}{0.1cm}
    \setlength{\belowcaptionskip}{-0.5cm}
    \centering
    \footnotesize
    \begin{tabular}{p{15cm}}
    \toprule
        \textbf{Retrieval Prompting for HotpotQA} \\
        Answer the given question in JSON format, you can refer to the document provided. \\
        As an assistant, your task is to answer the question based on the given knowledge. Your answer should be after <Answer> in JSON format with key "answer" and its value should be string. \\
        The given knowledge will be embraced by <doc> and </doc> tags. You can refer to the knowledge to answer the question. If the knowledge does not contain the answer, answer the question directly. \\
        
        There are some examples for you to refer to: \\
        <doc> \\
        \{\{KNOWLEDGE FOR YOUR REFERENCE\}\} \\
        </doc> \\
        <Question>: What is the name of this American musician, singer, actor, comedian, and songwriter, who worked with Modern Records and born in December 5, 1932? \\
        <Answer>: \\
        \texttt{\`}\texttt{\`}\texttt{\`} json \\
        \{\{"answer": "Little Richard"\}\} \\
        \texttt{\`}\texttt{\`}\texttt{\`} \\
        
        <doc> \\
        \{\{KNOWLEDGE FOR YOUR REFERENCE\}\} \\
        </doc> \\
        <Question>: Between Chinua Achebe and Rachel Carson, who had more diverse jobs? \\
        <Answer>: \\
        \texttt{\`}\texttt{\`}\texttt{\`} json \\
        \{\{"answer": "Chinua Achebe"\}\} \\
        \texttt{\`}\texttt{\`}\texttt{\`} \\
        
        <doc> \\
        \{\{KNOWLEDGE FOR YOUR REFERENCE\}\} \\
        </doc> \\
        <Question>: Remember Me Ballin' is a CD single by Indo G that features an American rapper born in what year? \\
        <Answer>: \\
        \texttt{\`}\texttt{\`}\texttt{\`} json \\
        \{\{"answer": "1979"\}\} \\
        \texttt{\`}\texttt{\`}\texttt{\`} \\
        
        Now your question and reference knowledge are as follows. \\
        <doc> \\
        \{knowledge\} \\
        </doc> \\
        <Question>: \{question\} \\
        <Answer>: \\
    \bottomrule
    \end{tabular}
    \caption{Detailed prompt for retrieval on HotpotQA}
    \label{tab:retrieve_prompt_hotpotQA}
\end{table*}

\begin{table*}[htbp]
    \setlength{\abovecaptionskip}{0.1cm}
    \setlength{\belowcaptionskip}{-0.5cm}
    \centering
    \footnotesize
    \begin{tabular}{p{15cm}}
    \toprule
        \textbf{Retrieval Prompting for MuSiQue} \\
        As an assistant, your task is to answer the question based on the given knowledge. Your answer should be after <Answer> in JSON format with key "answer" and its value should be string.  \\
        The given knowledge will be embraced by <doc> and </doc> tags. You can refer to the knowledge to answer the question. If the knowledge does not contain the answer, answer the question directly. \\
        
        There are some examples for you to refer to: \\
        
        <doc> \\
        \{\{KNOWLEDGE FOR YOUR REFERENCE\}\} \\
        </doc> \\
        <Question>: In which year did the publisher of In Cold Blood form? \\
        <Answer>: \\
        \texttt{\`}\texttt{\`}\texttt{\`} json \\
        \{\{"answer": "2001"\}\} \\
        \texttt{\`}\texttt{\`}\texttt{\`} \\
        
        <doc> \\
        \{\{KNOWLEDGE FOR YOUR REFERENCE\}\} \\
        </doc> \\
        <Question>: Who was in charge of the city where The Killing of a Sacred Deer was filmed? \\
        <Answer>: \\
        \texttt{\`}\texttt{\`}\texttt{\`} json \\
        \{\{"answer": "John Cranley"\}\} \\
        \texttt{\`}\texttt{\`}\texttt{\`} \\
        
        <doc> \\
        \{\{KNOWLEDGE FOR YOUR REFERENCE\}\} \\
        </doc> \\
        <Question>: Where on the Avalon Peninsula is the city that Signal Hill overlooks? \\
        <Answer>: \\
        \texttt{\`}\texttt{\`}\texttt{\`} json \\
        \{\{"answer": "eastern tip"\}\} \\
        \texttt{\`}\texttt{\`}\texttt{\`} \\
        
        Now your question and reference knowledge are as follows. \\
        <doc> \\
        \{knowledge\} \\
        </doc> \\
        <Question>: \{question\} \\
        <Answer>: \\
    \bottomrule
    \end{tabular}
    \caption{Detailed prompt for retrieval on MuSiQue}
    \label{tab:retrieve_prompt_musique}
\end{table*}

\begin{table*}[htbp]
    \setlength{\abovecaptionskip}{0.1cm}
    \setlength{\belowcaptionskip}{-0.5cm}
    \centering
    \footnotesize
    \begin{tabular}{p{15cm}}
    \toprule
        \textbf{Retrieval Prompting for 2Wiki-MultihopQA} \\
        As an assistant, your task is to answer the question based on the given knowledge. Your answer should be after <Answer> in JSON format with key "answer" and its value should be string. \\
        The given knowledge will be embraced by <doc> and </doc> tags. You can refer to the knowledge to answer the question. If the knowledge does not contain the answer, answer the question directly. \\
        
        There are some examples for you to refer to: \\
        <doc> \\
        \{\{KNOWLEDGE FOR YOUR REFERENCE\}\} \\
        </doc> \\
        <Question>: Which film came out first, Blind Shaft or The Mask Of Fu Manchu? \\
        <Answer>: \\
        \texttt{\`}\texttt{\`}\texttt{\`} json \\
        \{\{"answer": "The Mask Of Fu Manchu"\}\} \\
        \texttt{\`}\texttt{\`}\texttt{\`} \\
        
        <doc> \\
        \{\{KNOWLEDGE FOR YOUR REFERENCE\}\} \\
        </doc> \\
        <Question>: When did John V, Prince Of Anhalt-Zerbst's father die? \\
        <Answer>: \\
        \texttt{\`}\texttt{\`}\texttt{\`} json \\
        \{\{"answer": "12 June 1516"\}\} \\
        \texttt{\`}\texttt{\`}\texttt{\`} \\
        
        <doc> \\
        \{\{KNOWLEDGE FOR YOUR REFERENCE\}\} \\
        </doc> \\
        <Question>: Which film has the director who was born later, El Extrano Viaje or Love In Pawn? \\
        <Answer>: \\
        \texttt{\`}\texttt{\`}\texttt{\`} json \\
        \{\{"answer": "El Extrano Viaje"\}\} \\
        \texttt{\`}\texttt{\`}\texttt{\`} \\
        
        Now your question and reference knowledge are as follows. \\
        <doc> \\
        \{knowledge\} \\
        </doc> \\
        <Question>: \{question\} \\
        <Answer>: \\
    \bottomrule
    \end{tabular}
    \caption{Detailed prompt for retrieval on 2Wiki-MultihopQA}
    \label{tab:retrieve_prompt_2WikiMQA}
\end{table*}

\begin{table*}[htbp]
    \setlength{\abovecaptionskip}{0.1cm}
    \setlength{\belowcaptionskip}{-0.5cm}
    \centering
    \footnotesize
    \begin{tabular}{p{15cm}}
    \toprule
        \textbf{Iter-RetGen Prompting for HotpotQA} \\
        You should think step by step and answer the question after <Question> based on given knowledge embraced with <doc> and </doc>. Your answer should be after <Answer> in JSON format with key "thought" and "answer", their value should be string. \\

        Here are some examples for you to refer to: \\
        <doc> \\
        \{\{KNOWLEDGE FOR THE QUESTION\}\} \\
        </doc> \\
        <Question>: What is the name of this American musician, singer, actor, comedian, and songwriter, who worked with Modern Records and born in December 5, 1932? \\
        Let's think step by step. \\
        <Answer>: \\
        \texttt{\`}\texttt{\`}\texttt{\`} json \\
        \{\{
            "thought": "Artists who worked with Modern Records include Etta James, Joe Houston, Little Richard, Ike and Tina Turner and John Lee Hooker in the 1950s and 1960s. Of these Little Richard, born in December 5, 1932, was an American musician, singer, actor, comedian, and songwriter.",
            "answer": "Little Richard"
        \}\} \\
        \texttt{\`}\texttt{\`}\texttt{\`} \\
        
        <doc> \\
        \{\{KNOWLEDGE FOR THE QUESTION\}\} \\
        </doc> \\
        <Question>: Between Chinua Achebe and Rachel Carson, who had more diverse jobs? \\
        <Answer>: \\
        \texttt{\`}\texttt{\`}\texttt{\`} json \\
        \{\{
            "thought": "Chinua Achebe was a Nigerian novelist, poet, professor, and critic. Rachel Carson was an American marine biologist, author, and conservationist. So Chinua Achebe had 4 jobs, while Rachel Carson had 3 jobs. Chinua Achebe had more diverse jobs than Rachel Carson.",
            "answer": "Chinua Achebe"
        \}\} \\
        \texttt{\`}\texttt{\`}\texttt{\`} \\
        
        <doc> \\
        \{\{KNOWLEDGE FOR THE QUESTION\}\} \\
        </doc> \\
        <Question>: Remember Me Ballin' is a CD single by Indo G that features an American rapper born in what year? \\
        <Answer>: \\
        \texttt{\`}\texttt{\`}\texttt{\`} json \\
        \{\{
            "thought": "Remember Me Ballin' is the CD single by Indo G featuring Gangsta Boo. Gangsta Boo is Lola Mitchell's stage name, who was born in August 7, 1979, and is an American rapper.",
            "answer": "1979"
        \}\} \\
        \texttt{\`}\texttt{\`}\texttt{\`} \\
        
        Now based on the given doc, answer the question after <Question>. \\
        <doc> \\
        \{documents\} \\
        </doc> \\
        <Question>: \{question\} \\
        Let's think step by step. \\
        <Answer>: \\
    \bottomrule
    \end{tabular}
    \caption{Detailed prompt for Iter-RetGen on HotpotQA}
    \label{tab:iter-retgen_prompt_hotpotQA}
\end{table*}

\begin{table*}[htbp]
    \setlength{\abovecaptionskip}{0.1cm}
    \setlength{\belowcaptionskip}{-0.5cm}
    \centering
    \footnotesize
    \begin{tabular}{p{15cm}}
    \toprule
        \textbf{Iter-RetGen Prompting for MuSiQue} \\
        You should think step by step and answer the question after <Question> based on given knowledge embraced with <doc> and </doc>. Your answer should be after <Answer> in JSON format with key "thought" and "answer", their value should be string. \\

        Here are some examples for you to refer to: \\
        <doc> \\
        \{\{KNOWLEDGE FOR THE QUESTION\}\} \\
        </doc> \\
        <Question>: In which year did the publisher of In Cold Blood form? \\
        Let's think step by step. \\
        <Answer>: \\
        \texttt{\`}\texttt{\`}\texttt{\`} json \\
        \{\{
            "thought": "In Cold Blood was first published in book form by Random House. Random House was form in 2001.",
            "answer": "2011"
        \}\} \\
        \texttt{\`}\texttt{\`}\texttt{\`} \\
        
        <doc> \\
        \{\{KNOWLEDGE FOR THE QUESTION\}\} \\
        </doc> \\
        <Question>: Who was in charge of the city where The Killing of a Sacred Deer was filmed? \\
        Let's think step by step. \\
        <Answer>: \\
        \texttt{\`}\texttt{\`}\texttt{\`} json \\
        \{\{
            "thought": "The Killing of a Sacred Deer was filmed in Cincinnati. The present Mayor of Cincinnati is John Cranley. Therefore, John Cranley is in charge of the city.",
            "answer": "John Cranley"
        \}\} \\
        \texttt{\`}\texttt{\`}\texttt{\`} \\
        
        <doc> \\
        \{\{KNOWLEDGE FOR THE QUESTION\}\} \\
        </doc> \\
        <Question>: Where on the Avalon Peninsula is the city that Signal Hill overlooks? \\
        Let's think step by step. \\
        <Answer>: \\
        \texttt{\`}\texttt{\`}\texttt{\`} json \\
        \{\{
            "thought": "Signal Hill is a hill which overlooks the city of St. John's. St. John's is located on the eastern tip of the Avalon Peninsula.",
            "answer": "eastern tip"
        \}\} \\
        \texttt{\`}\texttt{\`}\texttt{\`} \\
        
        Now based on the given doc, answer the question after <Question>. \\
        <doc> \\
        \{documents\} \\
        </doc> \\
        <Question>: \{question\} \\
        Let's think step by step. \\
        <Answer>: \\
    \bottomrule
    \end{tabular}
    \caption{Detailed prompt for Iter-RetGen on MuSiQue}
    \label{tab:iter-retgen_prompt_musique}
\end{table*}

\begin{table*}[htbp]
    \setlength{\abovecaptionskip}{0.1cm}
    \setlength{\belowcaptionskip}{-0.5cm}
    \centering
    \footnotesize
    \begin{tabular}{p{15cm}}
    \toprule
        \textbf{Iter-RetGen Prompting for 2Wiki-MultihopQA} \\
        You should think step by step and answer the question after <Question> based on given knowledge embraced with <doc> and </doc>. Your answer should be after <Answer> in JSON format with key "thought" and "answer", their value should be string. \\

        Here are some examples for you to refer to: \\
        <doc> \\
        \{\{KNOWLEDGE FOR THE QUESTION\}\} \\
        </doc> \\
        <Question>: Which film came out first, Blind Shaft or The Mask Of Fu Manchu? \\
        Let's think step by step. \\
        <Answer>: \\
        \texttt{\`}\texttt{\`}\texttt{\`} json \\
        \{\{
            "thought": "Blind Shaft is a 2003 film, while The Mask Of Fu Manchu opened in New York on December 2, 1932. 2003 comes after 1932. Therefore, The Mask Of Fu Manchu came out earlier than Blind Shaft.",
            "answer": "The Mask Of Fu Manchu"
        \}\} \\
        \texttt{\`}\texttt{\`}\texttt{\`} \\
        
        <doc> \\
        \{\{KNOWLEDGE FOR THE QUESTION\}\} \\
        </doc> \\
        <Question>: When did John V, Prince Of Anhalt-Zerbst's father die? \\
        Let's think step by step. \\
        <Answer>: \\
        \texttt{\`}\texttt{\`}\texttt{\`} json \\
        \{\{
            "thought": "John V, Prince Of Anhalt-Zerbst was the son of Ernest I, Prince of Anhalt-Dessau. Ernest I, Prince of Anhalt-Dessau died on 12 June 1516.",
            "answer": "12 June 1516"
        \}\} \\
        \texttt{\`}\texttt{\`}\texttt{\`} \\
        
        <doc> \\
        \{\{KNOWLEDGE FOR THE QUESTION\}\} \\
        </doc> \\
        <Question>: Which film has the director who was born later, El Extrano Viaje or Love In Pawn? \\
        Let's think step by step. \\
        <Answer>: \\
        \texttt{\`}\texttt{\`}\texttt{\`} json \\
        \{\{
            "thought": "The director of El Extrano Viaje is Fernando Fernan Gomez, who was born on 28 August 1921. The director of Love In Pawn is Charles Saunders, who was born on 8 April 1904. 28 August 1921 comes after 8 April 1904. Therefore, Fernando Fernan Gomez was born later than Charles Saunders.",
            "answer": "El Extrano Viaje"
        \}\} \\
        \texttt{\`}\texttt{\`}\texttt{\`} \\
        
        Now based on the given doc, answer the question after <Question> \\
        <doc> \\
        \{documents\} \\
        </doc> \\
        <Question>: \{question\} \\
        Let's think step by step. \\
        <Answer>: \\
    \bottomrule
    \end{tabular}
    \caption{Detailed prompt for Iter-RetGen on 2Wiki-MultihopQA}
    \label{tab:iter-retgen_prompt_2WikiMQA}
\end{table*}

\begin{table*}[htbp]
    \setlength{\abovecaptionskip}{0.1cm}
    \setlength{\belowcaptionskip}{-0.5cm}
    \centering
    \footnotesize
    \begin{tabular}{p{15cm}}
    \toprule
        \textbf{Self-ask Prompting for HotpotQA} \\
        Solve the question with the given knowledge. \\
        Each line should start with either "Intermediate answer:", "Follow up:", "So the final answer is:", or "Are follow up questions needed here:". \\
        \# \\
        Question: What is the name of this American musician, singer, actor, comedian, and songwriter, who worked with Modern Records and born in December 5, 1932? \\
        Are follow up questions needed here: Yes. \\
        Follow up: Who worked with Modern Records? \\
        Intermediate answer: Artists worked with Modern Records include Etta James, Little Richard, Joe Houston, Ike and Tina Turner and John Lee Hooker. \\
        Follow up: Is Etta James an American musician, singer, actor, comedian, and songwriter, and was born in December 5, 1932? \\
        Intermediate answer: Etta James was born in January 25, 1938, not December 5, 1932, so the answer is no. \\
        Follow up: Is Little Richard an American musician, singer, actor, comedian, and songwriter, and was born in December 5, 1932? \\
        Intermediate answer: Yes, Little Richard, born in December 5, 1932, is an American musician, singer, actor, comedian and songwriter. \\
        So the final answer is: Little Richard \\
        \# \\
        Question: Between Chinua Achebe and Rachel Carson, who had more diverse jobs? \\
        Are follow up questions needed here: Yes. \\
        Follow up: What jobs did Chinua Achebe have? \\
        Intermediate answer: Chinua Achebe was a Nigerian (1) novelist, (2) poet, (3) professor, and (4) critic, so Chinua Achebe had 4 jobs. \\
        Follow up: What jobs did Rachel Carson have? \\
        Intermediate answer: Rachel Carson was an American (1) marine biologist, (2) author, and (3) conservationist, so Rachel Carson had 3 jobs. \\
        Follow up: Did Chinua Achebe have more jobs than Rachel Carson? \\
        Intermediate answer: Chinua Achebe had 4 jobs, while Rachel Carson had 3 jobs. 4 is greater than 3, so yes, Chinua Achebe had more jobs. \\
        So the final answer is: Chinua Achebe \\
        \# \\
        Question: Remember Me Ballin' is a CD single by Indo G that features an American rapper born in what year? \\
        Are follow up questions needed here: Yes. \\
        Follow up: Which American rapper is featured by Remember Me Ballin', a CD single by Indo G? \\
        Intermediate answer: Gangsta Boo \\
        Follow up: In which year was Gangsta Boo born? \\
        Intermediate answer: Gangsta Boo was born in August 7, 1979, so the answer is 1979. \\
        So the final answer is: 1979 \\
        \# \\
        Question: \{question\} \\
        Are follow up questions needed here: \\
    \bottomrule
    \end{tabular}
    \caption{Detailed prompt for self-ask on HotpotQA}
    \label{tab:self-ask_pormpt_hotpotQA}
\end{table*}

\begin{table*}[htbp]
    \setlength{\abovecaptionskip}{0.1cm}
    \setlength{\belowcaptionskip}{-0.5cm}
    \centering
    \footnotesize
    \begin{tabular}{p{15cm}}
    \toprule
        \textbf{Self-ask Prompting for MuSiQue} \\
        Solve the question with the given knowledge. \\
        Each line should start with either "Intermediate answer:", "Follow up:", "So the final answer is:", or "Are follow up questions needed here:". \\
        \# \\
        Question: In which year did the publisher of In Cold Blood form? \\
        Are follow up questions needed here: Yes. \\
        Follow up: What business published In Cold Blood? \\
        Intermediate answer: In Cold Blood was published in book form by Random House. \\
        Follow up: Which year witnessed the formation of Random House? \\
        Intermediate answer: Random House was form in 2001. \\
        So the final answer is: 2001 \\
        \# \\
        Question: Who was in charge of the city where The Killing of a Sacred Deer was filmed? \\
        Are follow up questions needed here: Yes. \\
        Follow up: In which city was The Killing of a Sacred Deer filmed \\
        Intermediate answer: The Killing of a Sacred Deer was filmed in Cincinnati. \\
        Follow up: Who was in charge of Cincinnati? \\
        Intermediate answer: The present Mayor of Cincinnati is John Cranley, so John Cranley is in charge. \\
        So the final answer is: John Cranley \\
        \# \\
        Question: Where on the Avalon Peninsula is the city that Signal Hill overlooks? \\
        Are follow up questions needed here: Yes. \\
        Follow up: What city does Signal Hill overlook? \\
        Intermediate answer: Signal Hill is a hill which overlooks the city of St. John's. \\
        Follow up: Where on the Avalon Peninsula is St. John's located? \\
        Intermediate answer: St. John's is located on the eastern tip of the Avalon Peninsula. \\
        So the final answer is: eastern tip \\
        \# \\
        Question: \{question\} \\
        Are follow up questions needed here: \\
    \bottomrule
    \end{tabular}
    \caption{Detailed prompt for self-ask on MuSiQue}
    \label{tab:self-ask_pormpt_musique}
\end{table*}

\begin{table*}[htbp]
    \setlength{\abovecaptionskip}{0.1cm}
    \setlength{\belowcaptionskip}{-0.5cm}
    \centering
    \footnotesize
    \begin{tabular}{p{15cm}}
    \toprule
        \textbf{Self-ask Prompting for 2Wiki-MultihopQA} \\
        Solve the question with the given knowledge. \\
        Each line should start with either "Intermediate answer:", "Follow up:", "So the final answer is:", or "Are follow up questions needed here:". \\
        Follow the examples below to answer the questions with natural language. \\
        \# \\
        Question: Which film came out first, Blind Shaft or The Mask Of Fu Manchu? \\
        Are follow up questions needed here: Yes. \\
        Follow up: When did Blind Shaft come out? \\
        Intermediate answer: Blind Shaft came out in 2003. \\
        Follow up: When did The Mask Of Fu Manchu come out? \\
        Intermediate answer: The Mask Of Fu Manchu came out in 1932. \\
        So the final answer is: The Mask Of Fu Manchu \\
        \# \\
        Question: When did John V, Prince Of Anhalt-Zerbst's father die? \\
        Are follow up questions needed here: Yes. \\
        Follow up: Who is the father of John V, Prince Of Anhalt-Zerbst? \\
        Intermediate answer: The father of John V, Prince Of Anhalt-Zerbst is Ernest I, Prince of Anhalt-Dessau. \\
        Follow up: When did Ernest I, Prince of Anhalt-Dessau die? \\
        Intermediate answer: Ernest I, Prince of Anhalt-Dessau died on 12 June 1516. \\
        So the final answer is: 12 June 1516 \\
        \# \\
        Question: Which film has the director who was born later, El Extrano Viaje or Love In Pawn? \\
        Are follow up questions needed here: Yes. \\
        Follow up: Who is the director of El Extrano Viaje? \\
        Intermediate answer: The director of El Extrano Viaje is Fernando Fernan Gomez. \\
        Follow up: Who is the director of Love in Pawn? \\
        Intermediate answer: The director of Love in Pawn is Charles Saunders. \\
        Follow up: When was Fernando Fernan Gomez born? \\
        Intermediate answer: Fernando Fernan Gomez was born on 28 August 1921. \\
        Follow up: When was Charles Saunders (director) born? \\
        Intermediate answer: Charles Saunders was born on 8 April 1904. \\
        So the final answer is: El Extrano Viaje \\
        \# \\
        Question: \{question\} \\
        Are follow up questions needed here: \\
    \bottomrule
    \end{tabular}
    \caption{Detailed prompt for self-ask on 2Wiki-MultihopQA}
    \label{tab:self-ask_pormpt_2WikiMQA}
\end{table*}